\documentclass[times,twocolumn,final]{elsarticle}

\usepackage{ipcai_arxiv}
\usepackage{framed,multirow}

\usepackage{amssymb}
\usepackage{latexsym}
\usepackage{url}
\usepackage{cite}
\usepackage{amsmath,amssymb,amsfonts,bm}
\usepackage{booktabs}
\usepackage{arydshln}
\usepackage{tabulary}
\usepackage{graphicx,epstopdf}
\usepackage{textcomp}
\usepackage{subcaption}
\usepackage[svgnames,table]{xcolor}
\usepackage{pifont}
\usepackage[colorlinks=true,citecolor=cyan,urlcolor=cyan]{hyperref}
\usepackage{placeins}
\usepackage{fancyhdr}
\usepackage{float}
\usepackage{listings}
\usepackage{makecell}
\usepackage{algorithm}
\usepackage{algorithmicx}
\usepackage{algpseudocode}
\usepackage[most]{tcolorbox}
\usepackage[utf8]{inputenc}
\usepackage{newunicodechar}
\newunicodechar{−}{\textminus}

\fancypagestyle{firstpagestyle}{
    \fancyhf{}
    
    \fancyhead{}
    \fancyfoot{}
    \fancyhead[CO]{\em \fontsize{9pt}{8pt}\selectfont This article has been accepted to the International Conference on Information Processing in Computer-Assisted Interventions, 2026.}
}

\begin{document}

\title{GEN-Guard: Correcting Generalization Failures for Deployable Federated Surgical AI}

\author[1,2]{Julia \snm{Alekseenko}\corref{cor}}
\author[2,3]{Pietro \snm{Mascagni}}
\author[3,4,5,6]{AI4SafeChole \snm{Consortium}}
\author[1,2]{Nicolas \snm{Padoy}}

\cortext[cor]{Corresponding author: alekseenko@unistra.fr}

\address[1]{University of Strasbourg, CNRS, INSERM, ICube, UMR7357, Strasbourg, France}
\address[2]{IHU Strasbourg, Strasbourg, France}
\address[3]{Bioimage Analysis Center, Fondazione Policlinico Universitario Agostino Gemelli IRCCS, Rome, Italy}
\address[4]{Azienda Ospedaliero-Universitaria Sant'Andrea, Rome, Italy}
\address[5]{Fondazione IRCCS Ca' Granda Ospedale Maggiore Policlinico di Milano, University of Milan, Milan, Italy}
\address[6]{Monaldi Hospital, AORN dei Colli, Naples, Italy}

\received{XXX}
\finalform{XXX}
\accepted{XXX}
\availableonline{XXX}
\communicated{XXX}

\begin{abstract}

\textbf{Purpose:} Federated Learning (FL) in surgical video AI enables collaborative model training without sharing sensitive data. However, standard evaluation practices -- selecting the ``best'' global model based only on validation data from participating hospitals -- can lead to suboptimal deployment choices. We identify this critical failure mode as \textit{performance leakage}, where the selected model overfits internal federation data and fails to generalize to unseen institutions, thereby \textit{undermining the core goal of FL: robust real-world generalization}.

\textbf{Method:} We propose GEN-Guard, a practical post-hoc framework to detect and correct generalization failures in federated surgical AI. It integrates Generalization Detection via Client-Blocked Evaluation (CBE), which validates performance on isolated client distributions to prevent performance leakage, and Generalization Correction through Disagreement-Aware Distillation (DAD), which learns adaptive feature-level corrections for cross-institutional robustness. Both components operate after standard FL convergence while providing robust support for zero-shot adaptation to unseen clinical environments.

\textbf{Results:} We first quantify the severity of performance leakage, observing \textbf{Model Selection Failures (MSFs) exceeding 80\%} under standard evaluation. GEN-Guard is evaluated on two multi-center clinical challenges: surgical phase recognition in laparoscopic cholecystectomy and polyp segmentation in colonoscopy. Across both datasets, GEN-Guard consistently corrects these failures, improving in-federation F1 scores by up to 2 points, unseen-institution performance by up to 3 points, and worst-case institutional performance by 3--9 points.

\textbf{Conclusions:} Performance leakage represents a systematic and previously under-recognized risk in federated surgical AI. GEN-Guard provides a practical, privacy-preserving solution for detecting and correcting such failures without altering federated training procedures. By improving cross-institutional robustness and zero-shot generalization, it strengthens the reliability of FL for real-world surgical deployment.
\end{abstract}

\maketitle
\thispagestyle{firstpagestyle}

\section{Introduction}\label{sec:intro}

Surgical video Artificial Intelligence (AI) analysis -- such as phase/step recognition \citep{lavanchy2024challenges}, anatomical segmentation \citep{murali2024cyclesam}, and complex decision-support tasks like Critical View of Safety (CVS) prediction \citep{mascagni2022multicentric} -- require large and diverse datasets for clinically reliable performance. However, the distributed and privacy-sensitive nature of surgical video data makes centralized training challenging. Traditional methods that aggregate data across institutions are often impractical due to strict sensitive data regulations \citep{conduah2025data}. Federated Learning (FL) \citep{mcmahan2017communication} enables each hospital to train models locally on its own data and share only model updates -- not the raw data -- with a central server, allowing data-rich collaborative learning while keeping patient data private.

\begin{figure*}[ht!]
    \centering
    \includegraphics[width=450px, trim={15mm 90mm 110mm 30mm}, clip]{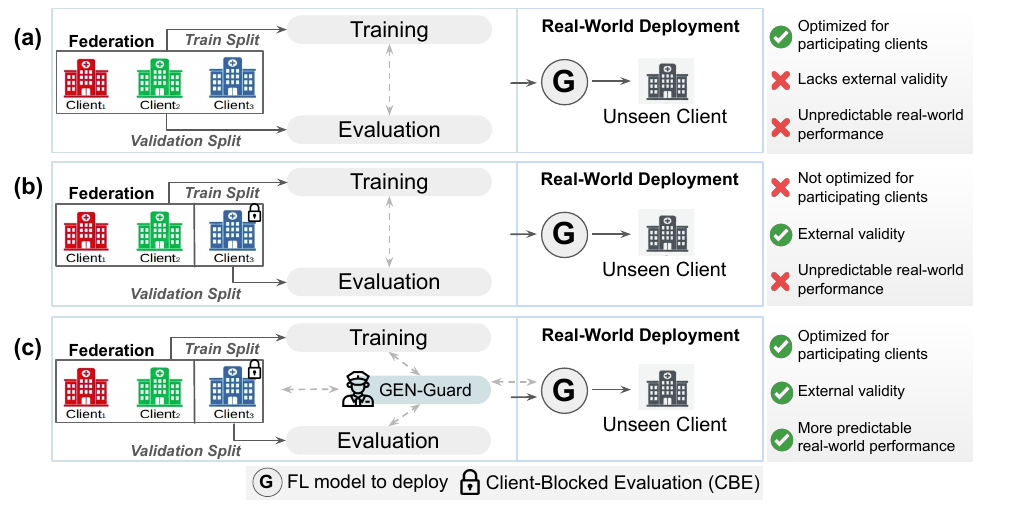}
    \caption{Training and deployment strategies in Federated Learning (FL). (a)~Conventional FL Evaluation. (b)~Baseline Client-Blocked Evaluation (CBE). (c)~\textbf{Proposed GEN-Guard Framework}.}
    \label{fig:into}
\end{figure*}

Recent studies show that FL can achieve performance comparable to centralized training across hospitals. For example, FL models have achieved strong results in surgical tasks, including segmentation ($Dice = 82.62_{\pm 0.09}$ vs.\ Centralized $83.21_{\pm 0.85}$ \citep{fang2026spatio}) and phase recognition ($F1 = 65.77_{\pm 0.89}$ vs.\ Centralized $69.29_{\pm 0.97}$ \citep{kassem2022federated}), demonstrating its effectiveness for decentralized surgical AI. \textit{Despite this promise, FL has yet to achieve its ultimate goal: robust generalization.} Work in \citep{yuan2022what} frames this through two dimensions: the \textit{Out-of-Sample Gap}, capturing drops on new data from known clients, and the more critical FL-specific \textit{Participation Gap}, where performance collapses on data from unseen client distributions. In surgical FL, this gap is evident: on external-site evaluation, FL performance drops sharply: phase recognition \citep{kassem2022federated} decreases by 23\% (65.77 F1 on federated clients $\to$ 42.91 F1 externally) and segmentation \citep{fang2026spatio} by 4.7\% (82.62 Dice $\to$ 77.97 Dice), with worst-case declines exceeding 65\% \citep{kirchner2025federated}. This highlights a key deployment bottleneck: \textit{FL-trained models often fail to generalize effectively across unseen clinical sites}. A systematic review \citep{teo2024federated} confirms that few FL studies in healthcare report real-world deployments or assess performance on unseen institutions.

This bottleneck arises from the multi-dimensional nature of the generalization gap in FL \citep{yuan2022what}. Conventional FL evaluations often measure performance only on participating clients, providing a limited view of true generalization. Two critical but overlooked factors contribute to this:
\begin{itemize}
    \item \textit{Institutional Variation} -- Differences in data distributions across hospitals due to equipment, workflows, and patient demographics.
    \item \textit{Deployment Diversity} -- The gap between training environments within the federation and the diverse, unseen conditions beyond the federation.
\end{itemize}
Most studies focus on the first factor while neglecting the second. We identify the root cause of this failure as \textit{Performance Leakage} in model selection: models appear stable on correlated validation in-federation data but fail on unseen institutions.

In summary, although FL is feasible for decentralized surgical video AI, reliable generalization remains challenging. To our knowledge, no prior work systematically quantifies Model Selection Failures (MSFs) in surgical video FL or corrects them. Existing methods frequently adopt evaluation protocols that introduce bias, causing generalization collapse across unseen sites. As a solution, \textbf{we propose GEN-Guard, a deployment-oriented, post-hoc framework} that operates on a discrete set of models after the federated training process. It detects and corrects MSFs via two components: Client-Blocked Evaluation (CBE) and Disagreement-Aware Distillation (DAD), promoting cross-hospital generalization of federated surgical AI without modifying or adding communication to the main FL training protocol.

\section{Related Work}

\subsection{Generalization and Domain Shift}

A central challenge in medical FL is robust generalization across heterogeneous clinical environments \citep{li2025challenges}. FL models often underperform on unseen sites due to severe non-IID data \citep{zhao2018federated}. For example, federated models show promise in surgical outcome prediction \citep{ren2025federated} or pre-operative analysis \citep{tzortzis2025towards} but decline on external hospitals, and greater site heterogeneity reduces segmentation accuracy \citep{luo2023influence}, \textit{highlighting the limits of current FL deployment frameworks}.

Methodological advances, including personalized FL, domain generalization, and meta-learning \citep{liu2025unified, zhang2023grace, khoee2024domain} -- aim to improve robustness by adapting global models to local data or learning domain-invariant representations. Data-centric strategies, i.e., diverse training datasets or synthetic augmentation \citep{rujas2024synthetic}, have also been explored. However, these approaches mainly enhance in-federation consistency and often overlook evaluation and model selection, where bias and information leakage remain.

Systematic reviews \citep{crowson2022systematic, teo2024federated} indicate that \textit{most FL studies in healthcare lack prospective external validation}. In surgical FL, strong institutional biases and persistent non-IID data further limit reliable deployment. In response, recent efforts call for standardized multi-institutional benchmarks \citep{karargyris2023federated} and open reporting of out-of-federation performance as essential steps toward deployable and generalizable surgical FL.

\subsection{Model Selection and Evaluation Bias}

Systematic external validation is essential to assess generalization to unseen clients. Models should be evaluated on independent, multi-center datasets excluded from training, providing realistic estimates of real-world performance \citep{yuan2022what, bujotzek2025real}.

\begin{figure*}[t]
    \centering
    \includegraphics[width=450px, trim={1.5cm 8.8cm 0 3.8cm}, clip]{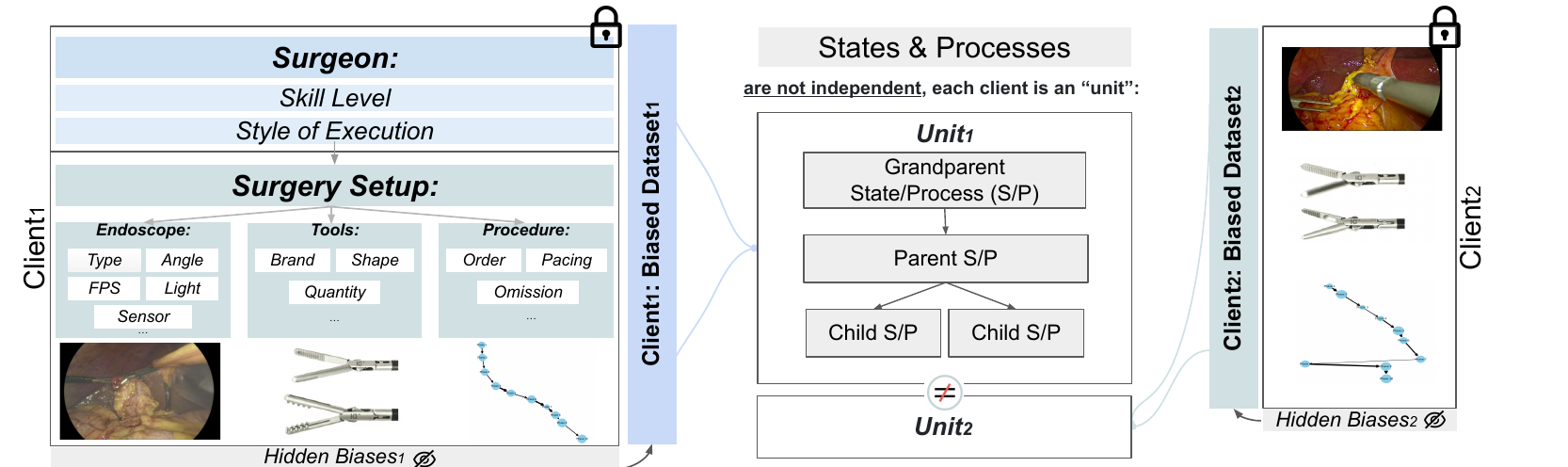}
    \caption{Sources of hidden biases in surgical video datasets and the ``Unit'' analogy.}
    \label{fig:client_bias}
\end{figure*}

In contrast, FL model selection typically relies on metrics from participating clients, introducing bias and causing \textit{Performance Leakage}. While FedAvg \citep{mcmahan2017communication} provides a baseline, methods like FedProx \citep{li2020federated} and SCAFFOLD \citep{karimireddy2020scaffold} handle optimization under client heterogeneity but do not ensure generalization or external validity. While CCVR \citep{luo2021no} proposes post-hoc calibration, it does not improve the learned representation space, is limited to classification tasks, and requires ground-truth labels. Fine-tuning like FedFTG \citep{zhang2022fine} adds a full training step and requires supervision, making it impractical for deployment. Both methods lack full external validity.

\section{Methodology}

\subsection{Problem Formulation}

Federated Learning (FL) aims to optimize a global parameter vector $\theta$ by minimizing the aggregate empirical risk across $K$ clients. The empirical risk ($\mathcal{L}(\theta)$) is defined directly over the model weights, where the objective is to find a global $\theta$ that performs optimally across all distributed datasets $\{\mathcal{D}_k\}_{k=1}^{K}$:

\begin{equation}
\min_{\theta} \mathcal{L}(\theta) = \sum_{k=1}^{K} \frac{n_k}{N} \mathcal{L}_k(\theta),
\quad
\mathcal{L}_k(\theta) = \frac{1}{n_k} \sum_{i \in \mathcal{D}_k} \ell(f_{\theta}(x_i), y_i),
\label{eq:fl_objective}
\end{equation}

\noindent where $n_k$ is the local data size at client $k$, $N=\sum_{k} n_k$ the total, and $\ell(\cdot)$ the sample loss.

The central server coordinates $T$ communication rounds. At each round $t$, the global model parameters are updated via weighted aggregation of the locally trained client models:

\begin{equation}
\theta^{t+1} = \sum_{k=1}^{K} \frac{n_k}{N} \theta_k^t,
\end{equation}

\noindent where $\theta_k^{t}$ denotes the local model parameters at client $k$, and $\theta^{t+1}$ represents the aggregated global model weights for the next round.

Institutional distribution heterogeneity ($\mathcal{D}_{k_1} \ne \mathcal{D}_{k_2}$) introduces a covariate generalization gap. Consequently, minimizing empirical risk across participating clients does not guarantee optimal generalization to unseen institutional distributions.

In standard FL practice, model selection is performed by choosing

\begin{equation}
\theta^{*} = \arg\max_{t \in \{1,\dots,T\}}
\mathbb{E}_{k \in \mathcal{C}_{\text{train}}}
\left[ \mathcal{G}(f_{\theta^{t}}, \mathcal{D}_k^{val}) \right],
\end{equation}

\noindent where $\mathcal{C}_{\text{train}}$ denotes participating clients and $\mathcal{D}_k^{val}$ their respective validation distributions.

We define \emph{performance leakage} as the condition where

\begin{equation}
\theta^{*} \neq
\arg\max_{t}
\mathbb{E}_{\mathcal{D}_{\text{unseen}}}
\left[ \mathcal{G}(f_{\theta^{t}}) \right],
\end{equation}

\noindent i.e., when the model selected using internal federation validation does not coincide with the model that would maximize performance on unseen institutional distributions. Performance leakage thus reflects a biased proxy of external generalization induced by federation-internal validation.

This mismatch results in a \emph{Model Selection Failure (MSF)} at deployment, where the chosen global model underperforms on previously unseen clients.

\textit{Goal:} Correct MSF by learning a deployment model that maximizes generalization to unseen clients. In this work, we define this model as $f_{\text{GEN-Guard}}$:

\begin{equation}
\max_{f_{\text{GEN-Guard}}}
\mathbb{E}_{\mathcal{D}_{\text{unseen}}}
[\mathcal{G}(f_{\text{GEN-Guard}})],
\end{equation}

\noindent where $\mathcal{G}$ denotes a task-specific performance metric evaluating the model on unseen institutional distributions.

\subsection{Hidden Biases in Federated Surgical Video Datasets}

Surgical video datasets are inherently non-homogeneous, shaped by institution-specific factors that create client-biased datasets in FL, where local features are highly interdependent and correlated \citep{eckhoff2023sages} (Figure~\ref{fig:client_bias}). As noted by a SAGES consensus \citep{eckhoff2023sages}, ignoring such dataset-specific biases can systematically over- or underestimate algorithmic performance. We conceptualize client-level heterogeneity using a hierarchical States/Processes (S/P) analogy, inspired by multi-level modeling in clinical and surgical studies \citep{eckhoff2023sages}:
\begin{itemize}
\item \textit{Grandparent S/P: Surgeon Characteristics:} Skill, style, and decision-making affect temporal dynamics, instrument trajectories, and scene composition.
\item \textit{Parent S/P: Surgical Setup}: Variations in endoscope settings, tool usage, and procedural execution introduce distinct visual and procedural patterns.
\item \textit{Child S/P: Video Data:} Captures the combined effects of Grandparent and Parent levels.
\end{itemize}
Together, these factors create a unique institutional fingerprint with intra-client feature correlations. This challenges the statistical independence assumptions required by conventional learning algorithms and FL. Conceptually, each client is a correlated data ``Unit''. Evaluating the model on a pooled validation set across these units encourages the model to evaluate on, and thus capture, the same client-specific patterns it just learned. This explains strong in-federation performance yet poor external robustness.

\subsection{GEN-Guard Framework}

Unlike traditional Federated Domain Generalization (FedDG) which integrates regularizers directly into the optimization objective, GEN-Guard is a post-hoc selection and refinement framework. It is integrated as:
\begin{itemize}
    \item \textbf{Standard FL Training:} The federation optimizes the empirical risk in Equation~\ref{eq:fl_objective}, producing a discrete trajectory of model checkpoints $\mathcal{S} = \{f_{\theta^1}, \dots, f_{\theta^T}\}$;
    \item \textbf{Generalization Detection \& Correction:} After training, GEN-Guard operates exclusively on the set $\mathcal{S}$ to identify and refine a deployable model.
\end{itemize}

\subsubsection{Detection: Client-Blocked Evaluation}

\begin{figure*}[t]
    \centering
    \includegraphics[width=450px, trim={20mm 120mm 10mm 30mm}, clip]{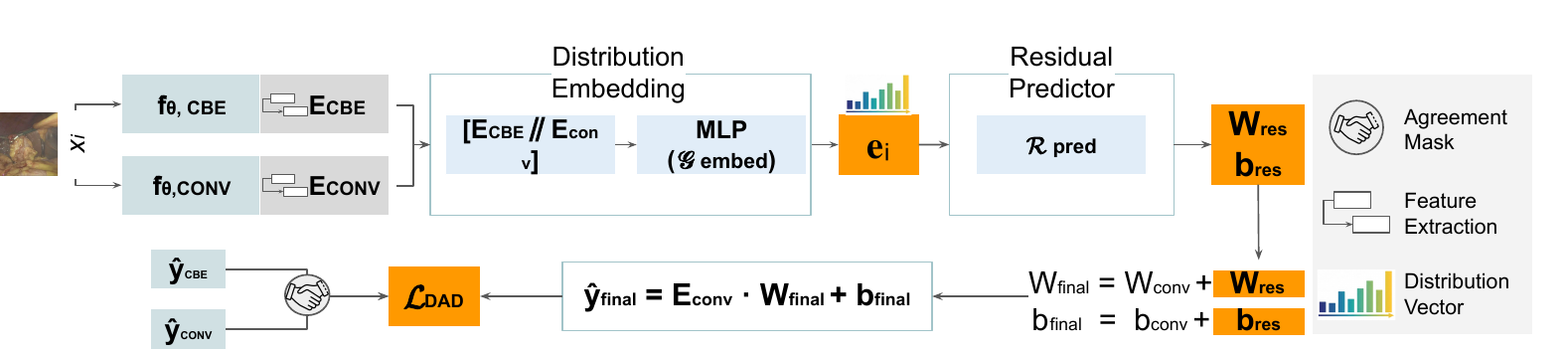}
    \caption{The Disagreement-Aware Distillation (DAD) pipeline.}
    \label{fig:DAD}
\end{figure*}

The Client-Blocked Evaluation (CBE) challenges the assumption that mixed-client validation predicts performance on unseen clients by enforcing client-level validation (see Figure~\ref{fig:into}).

Let $f_\theta$ be a model, $\mathcal{C}_{\text{train}}$ the set of training clients, $\mathcal{V}_{\text{mix}}$ a conventional mixed validation set, and $\mathcal{V}_{\text{block}, c}$ the client-isolated validation split for client $c$. The training set remains identical; only the validation split differs, and the isolated client used for validation is never included in training. Conventional FL selects the best model by maximizing a metric (e.g., F1) on the mixed set:
\begin{equation}
\hat{f}_{\theta, \text{CONV}} = \arg \max_{f_\theta} \text{Metric}(f_\theta, \mathcal{V}_{\text{mix}}).
\end{equation}

CBE identifies the optimal model by maximizing the robustness expectation on a held-out (blocked) client distribution, which serves as a proxy for the worst-case domain shift in the federation:

\begin{equation}
\hat{f}_{\theta, \text{CBE}} = \arg \max_{f_\theta} \mathbb{E}_{x \sim \mathcal{V}_{\text{block}, c}} \left[ \text{Metric}(f_\theta, x) \right].
\end{equation}

Here, $c \in \mathcal{C}_{\text{block}}$ is a client excluded from training. We define this evaluation as a worst-case generalization scenario because the model must perform ``blindly'' on a novel distribution without the benefit of site-specific artifacts (e.g., center-specific lighting or surgical tools) encountered during FL optimization. By maximizing performance on this unseen center, the framework prioritizes invariance, ensuring that the selected model remains robust to the domain shifts it is likely to face during deployment.

CBE tests whether conventional selection aligns with robust generalization: \textit{No Selection Failure:} $\hat{f}_{\theta, \text{CONV}} = \hat{f}_{\theta, \text{CBE}}$ -- mixed-validation evaluation is reliable; \textit{Model Selection Failure (MSF):} $\hat{f}_{\theta, \text{CONV}} \neq \hat{f}_{\theta, \text{CBE}}$ -- conventional evaluation is unreliable.

Thus, CBE provides the foundation for GEN-Guard, serving as the explicit \textit{Generalization Detection} mechanism that guides subsequent correction.

\subsubsection{Correction: Disagreement-Aware Distillation}

DAD transfers feature-level knowledge from the robust Client-Blocked model ($f_{\theta, \text{CBE}}$) to the conventional model ($f_{\theta, \text{CONV}}$), focusing on disagreements to correct biases and improve cross-institutional generalization (Figure~\ref{fig:DAD}).

For each sample $x_i$, normalized feature embeddings from the most abstract high-level layer are extracted from both models ($\mathbf{E}_{\text{CBE}}, \mathbf{E}_{\text{CONV}} \in \mathbb{R}^{\text{embed\_dim}}$). A small MLP, $\mathcal{G}_{\text{embed}}$, then produces a low-dimensional distribution embedding:

\begin{equation}
\mathbf{e}_i \in \mathbb{R}^{\text{dist\_dim}} = \mathcal{G}_{\text{embed}}([\mathbf{E}_{\text{CBE}} \parallel \mathbf{E}_{\text{CONV}}]),
\end{equation}
which feeds a residual predictor $\mathcal{R}_{\text{pred}}$ to generate residual weights and biases ($\mathbf{W}_{\text{res}}, \mathbf{b}_{\text{res}}$). Final predictions are corrected as:
\begin{equation}
\hat{\mathbf{y}}_{\text{final}} = \mathbf{E}_{\text{CONV}} \cdot (\mathbf{W}_{\text{CONV}} + \mathbf{W}_{\text{res}}) + (\mathbf{b}_{\text{CONV}} + \mathbf{b}_{\text{res}}).
\end{equation}

The DAD loss is a weighted sum of agreement and disagreement components:
\begin{equation}
\mathcal{L}_{\text{DAD}} = w_{\text{agree}} \cdot \mathcal{L}_{\text{agree}} + w_{\text{disagree}} \cdot \mathcal{L}_{\text{disagree}}.
\end{equation}

\textbf{Disagreement Loss:} Heavily weighted ($w_{\text{disagree}} > w_{\text{agree}}$) with higher temperature $T_{\text{disagree}}$, combining KL divergence ($\mathcal{L}_{\text{KL}}$) and cosine similarity $\mathcal{L}_{\text{COS}}$ on samples where $\hat{f}_{\theta, \text{CBE}}$ and $\hat{f}_{\theta, \text{CONV}}$ differ:
\begin{equation}
\mathcal{L}_{\text{disagree}} = \sum_{i \in \mathcal{M}_{\text{disagree}}} \left( \mathcal{L}_{\text{KL}}\!\left(\frac{\hat{\mathbf{y}}_{\text{final}, i}}{T_{\text{dis}}} \Big\| \frac{\hat{\mathbf{y}}_{\text{CBE}, i}}{T_{\text{dis}}}\right) + \mathcal{L}_{\text{COS}}(\hat{\mathbf{y}}_{\text{final}, i}, \hat{\mathbf{y}}_{\text{CBE}, i}) \right).
\end{equation}

\textbf{Agreement Loss:} Applied on samples where predictions align, with smaller weight and temperature $T_{\text{agree}}$:
\begin{equation}
\mathcal{L}_{\text{agree}} = \sum_{i \in \mathcal{M}_{\text{agree}}} \mathcal{L}_{\text{KL}}\!\left(\frac{\hat{\mathbf{y}}_{\text{final}, i}}{T_{\text{agree}}} \Big\| \frac{\hat{\mathbf{y}}_{\text{CBE}, i}}{T_{\text{agree}}}\right).
\end{equation}

For in-federation clients, an optional supervised loss ($\mathcal{L}_{\text{supervised}}$) ensures task accuracy. On unseen clients, DAD performs zero-shot without $\mathcal{L}_{\text{supervised}}$ by default.

\subsection{Convergence and Complexity Analysis}

GEN-Guard adds minimal overhead while correcting generalization bias:
\begin{itemize}
\renewcommand{\labelitemi}{--}
\item \textit{FL Training:} The global model ($f_\theta$) converges under the base FL algorithm (e.g., FedAvg), with GEN-Guard remaining post-hoc and optimizer-agnostic, preserving the base algorithm's convergence behavior.
\item \textit{DAD:} A shallow refinement process applied to selected checkpoints for a few epochs ($T_{\max}$). This pre-deployment step avoids the iterative FL communication rounds typical of federated updates.
\end{itemize}

Overall, GEN-Guard preserves base FL efficiency, especially for compute-intensive video models (Table~\ref{tab:complexity}).

\begin{table}[h]
\centering
\caption{Complexity comparison of GEN-Guard and standard FL.}
\label{tab:complexity}
\resizebox{\columnwidth}{!}{%
\begin{tabular}{lcc}
\toprule
\textbf{Metric} & \textbf{Standard FL} & \textbf{GEN-Guard} \\
\midrule
Training Communication ($T$ rounds) & $O(N|W|)$ & -- (no additional rounds) \\
Detection Overhead (CBE) & $O(|\mathcal{V}_{\text{mix}}|)$ & $O(|\mathcal{V}_{\text{block}}|)$ (extra fwd.) \\
Correction Overhead (DAD) & -- & $O(T_{\max}|\mathcal{V}_{\text{local}}|C_{\mathcal{G},\mathcal{R}})$ \\
Inference Overhead (DAD) & $O(C_{\text{CNN}})$ & $\approx O(C_{\text{CNN}})$ \\
\bottomrule
\end{tabular}%
}
\footnotesize{$N$ = number of clients; $|\mathbf{W}|$ = model size; $C_{\text{CNN}}$ = CNN forward pass cost; $C_{\mathcal{G},\mathcal{R}}$ = parameter generator cost ($C_{\mathcal{G},\mathcal{R}} \ll C_{\text{CNN}}$).}
\end{table}

\section{Results}\label{sec:results}

\subsection{Experimental Evaluation}

We evaluate GEN-Guard on two multi-institutional surgical video datasets for distinct tasks. \textit{Multi-Cholec (MultiChole2022)} \citep{kassem2022federated} targets Laparoscopic Cholecystectomy Phase Recognition (6 phases) with 180 de-identified videos from 5 centers (A--E) via the MOSaiC platform \citep{mazellier2023mosaic}: Gemelli, Sant'Andrea, Ca'~Granda, Mondaldi (25 videos each), and Cholec80 (80 videos). We used the same dataset split strategy as in \citep{kassem2022federated}: for Cholec80, 40 videos for training, 8 for validation, and 32 for testing; for the other four MultiChole2022 datasets, 13 videos for training, 6 for validation, and 6 for testing. \textit{PolypGen} \citep{ali2023multi} is used for Polyp Segmentation, containing 1,537 images from 6 centers (A--F) across Norway, France, UK, Egypt, and Italy (x2). The dataset was partitioned into training, validation, and testing sets using a randomized stratified split following a 70-15-15\% ratio.

We use repeated cross-validation, partitioning each dataset into three sets: \textit{In-Federation (In-Fed)} for training FL models; \textit{Held-out (Blocked)} for Client-Blocked Evaluation (CBE) as the generalization detection mechanism; and \textit{Out-of-Federation (Out-Fed)} for testing zero-shot generalization on a never-seen client.

Experiments focused on clients with extreme data imbalances, showing the strongest institutional bias: in \textit{Multi-Cholec}, the smallest and largest Out-Fed clients are E and B, respectively; in \textit{PolypGen}, they are F and C.

We adopt the model architectures and pre-processing from the original papers \citep{kassem2022federated, ali2023multi}. We evaluate three FL algorithms: FedAvg \citep{mcmahan2017communication} as baseline, FedProx ($\mu = 0.2$) \citep{li2020federated} to mitigate client drift on Non-IID data, and SCAFFOLD \citep{karimireddy2020scaffold} for variance-reduced convergence. FL training used 10 global rounds with local epochs and optimizers as in the original datasets. GEN-Guard is trained for up to $T_{\max}=5$ with early stopping, using Adam ($\text{lr} = 10^{-4}$) and hyperparameters $T_{\text{disagree}} = w_{\text{disagree}} = 2.0$, $T_{\text{agree}} = w_{\text{agree}} = 1.0$. Both $\mathcal{G}_{\text{embed}}$ and $\mathcal{R}_{\text{pred}}$ have 2 MLP layers for efficiency. We train models on Nvidia V100 GPUs, using Flower FL research framework \citep{beutel2020flower}.

\subsection{Generalization Detection: Model Selection Failure Results}

\begin{figure*}[t]
    \centering
    \includegraphics[width=450px]{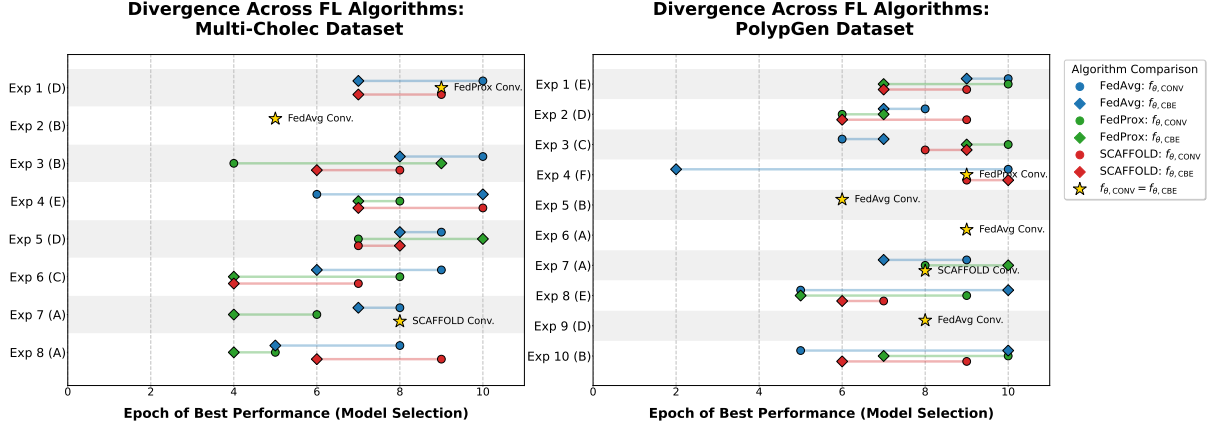}
    \caption{Divergence of FL algorithms: generalization detection via Client-Blocked Evaluation (CBE) across different clinical datasets. Each row represents an experimental setup, where the letter in parentheses denotes the held-out client (e.g., ``Exp~1~(D)'' indicates Client D was used for evaluation.)}
    \label{fig:combined_divergence}
\end{figure*}

Figure~\ref{fig:combined_divergence} shows the divergence between the conventionally selected model ($f_{\theta, \text{CONV}}$) and the robust CBE-selected model ($f_{\theta, \text{CBE}}$). A Model Selection Failure (MSF) occurs when $\hat f_{\theta, \text{CONV}} \neq \hat f_{\theta, \text{CBE}}$.

\begin{table}[h]
\centering
\caption{MSF across datasets and FL algorithms; runs = experiments.}
\label{tab:msf}
\resizebox{\columnwidth}{!}{%
\begin{tabular}{l c c c | l c c c}
\hline
\multicolumn{4}{c|}{\textbf{Multi-Cholec}} & \multicolumn{4}{c}{\textbf{PolypGen}} \\
\textbf{Algorithm} & \textbf{Runs} & \textbf{MSF} & \textbf{MSF \%} &
\textbf{Algorithm} & \textbf{Runs} & \textbf{MSF} & \textbf{MSF \%} \\
\hline
FedAvg   & 8 & 7 & 87.5 & FedAvg   & 10 & 7 & 70.0 \\
FedProx  & 7 & 6 & 85.7 & FedProx  & 7  & 6 & 85.7 \\
SCAFFOLD & 7 & 6 & 85.7 & SCAFFOLD & 7  & 6 & 85.7 \\
\hline
\multicolumn{8}{c}{\textbf{Total: 46 runs, MSF: 38, Overall: 82.6\%}} \\
\hline
\end{tabular}%
}
\end{table}

Across 46 experiments, MSF occurred in 38 cases (82.6\%), as shown in Table~\ref{tab:msf}, confirming that conventional model selection is often unreliable. FedProx and SCAFFOLD runs were skipped when FedAvg had already converged to avoid redundancy.

Additional statistics offer further insight:
\begin{itemize}
    \item \textit{Mean Absolute Divergence (2.6 Rounds):} On average, the selected model is 2.6 global rounds away from the optimal model for generalization.
    \item \textit{Maximum Selection Risk (8 Rounds):} In the worst-case, conventional selection picks a model 8 epochs away from the optimal, posing a severe deployment risk.
    \item \textit{Bias Trend (27 Positive / 11 Negative):} The influence of positive values indicates overfitting to institutional signatures within the known federation data.
\end{itemize}

\subsection{Generalization Correction Results}

Table~\ref{tab:average_gains_summary_full} shows that GEN-Guard consistently improves Multi-Cholec F1 scores across all FL protocols, boosting average performance (up to 2.8 points) and reducing variance. In worst-case scenarios, GEN-Guard improves the weakest models (up to 8.9 F1 points), demonstrating its ability to safeguard against catastrophic failures.

\begin{table*}[h]
\centering
\caption{GEN-Guard performance on Multi-Cholec: average F1$_{\pm\text{std}}$ and worst-case correction.}
\label{tab:average_gains_summary_full}
\resizebox{400px}{!}{%
\begin{tabular}{lcccccc}
\toprule
\textbf{FL Protocol} & \multicolumn{2}{c}{\textbf{In-Fed F1 (Supervised)}} & \multicolumn{2}{c}{\textbf{Held-Out F1 (Zero-shot)}} & \multicolumn{2}{c}{\textbf{Out-Fed F1 (Zero-shot)}} \\
\cmidrule(lr){2-3} \cmidrule(lr){4-5} \cmidrule(lr){6-7}
& $\mathbf{F1}_{\text{CONV}}$ & $\mathbf{F1}_{\text{GEN-Guard}}$ ($\mathbf{\Delta}$) & $\mathbf{F1}_{\text{CONV}}$ & $\mathbf{F1}_{\text{GEN-Guard}}$ ($\mathbf{\Delta}$) & $\mathbf{F1}_{\text{CONV}}$ & $\mathbf{F1}_{\text{GEN-Guard}}$ ($\mathbf{\Delta}$) \\
\midrule
FedAvg   & $69.53_{\pm2.14}$ & $\mathbf{71.47_{\pm1.90}}$ ($\mathbf{\uparrow 1.94}$) & $64.57_{\pm2.99}$ & $\mathbf{66.89_{\pm4.27}}$ ($\mathbf{\uparrow 2.32}$) & $65.82_{\pm2.47}$ & $\mathbf{67.18_{\pm1.40}}$ ($\mathbf{\uparrow 1.36}$) \\
FedProx  & $65.58_{\pm3.26}$ & $\mathbf{67.31_{\pm1.48}}$ ($\mathbf{\uparrow 1.73}$) & $64.96_{\pm4.35}$ & $\mathbf{67.72_{\pm3.82}}$ ($\mathbf{\uparrow 2.76}$) & $62.12_{\pm3.70}$ & $\mathbf{64.27_{\pm1.20}}$ ($\mathbf{\uparrow 2.15}$) \\
SCAFFOLD & $68.98_{\pm2.68}$ & $\mathbf{71.25_{\pm2.60}}$ ($\mathbf{\uparrow 2.27}$) & $66.45_{\pm1.68}$ & $\mathbf{68.79_{\pm1.86}}$ ($\mathbf{\uparrow 2.34}$) & $66.04_{\pm2.88}$ & $\mathbf{67.72_{\pm1.60}}$ ($\mathbf{\uparrow 1.68}$) \\
\midrule
\multicolumn{7}{l}{\textbf{Worst-Case F1 Correction}} \\
\midrule
FedAvg   & $65.71$ & $\mathbf{67.86}$ ($\mathbf{\uparrow 2.15}$) & $62.08$ & $\mathbf{67.49}$ ($\mathbf{\uparrow 5.41}$) & $62.46$ & $\mathbf{65.51}$ ($\mathbf{\uparrow 3.05}$) \\
FedProx  & $59.89$ & $\mathbf{65.95}$ ($\mathbf{\uparrow 6.06}$) & $55.47$ & $\mathbf{62.36}$ ($\mathbf{\uparrow 6.88}$) & $54.57$ & $\mathbf{63.50}$ ($\mathbf{\uparrow 8.93}$) \\
SCAFFOLD & $66.19$ & $\mathbf{70.97}$ ($\mathbf{\uparrow 4.78}$) & $66.44$ & $\mathbf{70.75}$ ($\mathbf{\uparrow 4.31}$) & $62.49$ & $\mathbf{66.53}$ ($\mathbf{\uparrow 4.04}$) \\
\bottomrule
\end{tabular}%
}
\end{table*}

\begin{figure}[t]
    \centering
    \begin{subfigure}{\linewidth}
        \centering
        \includegraphics[width=1\linewidth, trim={0 87mm 0 65mm}, clip]{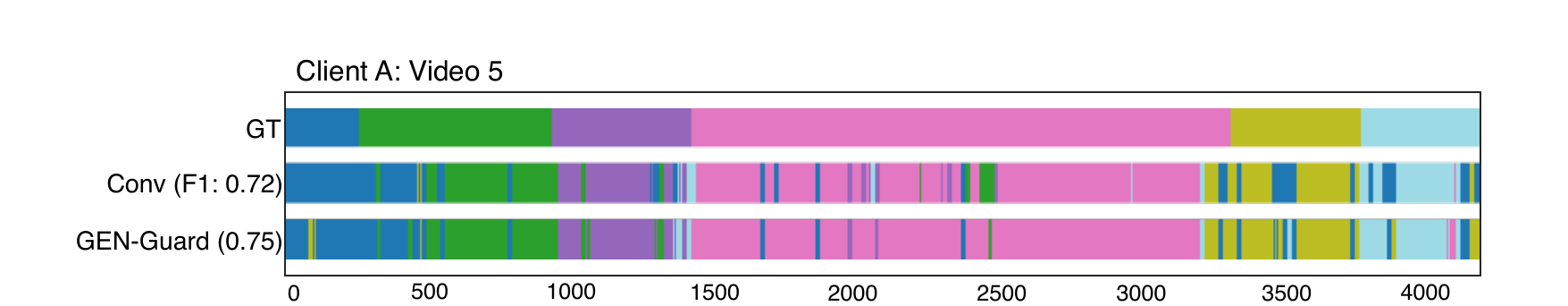}
        \vspace{-7.5mm}
    \end{subfigure}
    \begin{subfigure}{\linewidth}
        \centering
        \includegraphics[width=1\linewidth, trim={0 87mm 0mm 65mm}, clip]{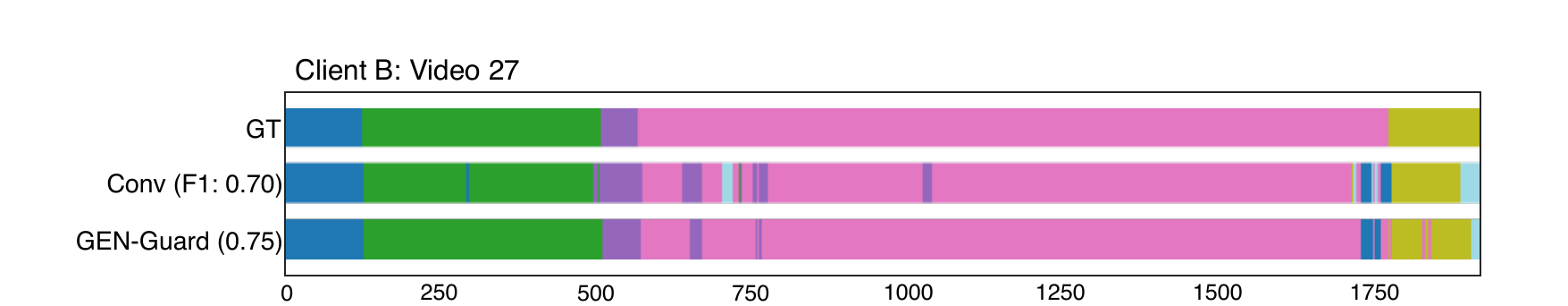}
        \vspace{-5.0mm}
    \end{subfigure}
    \begin{subfigure}{\linewidth}
        \centering
        \includegraphics[width=1\linewidth, trim={0 87mm 0 70mm}, clip]{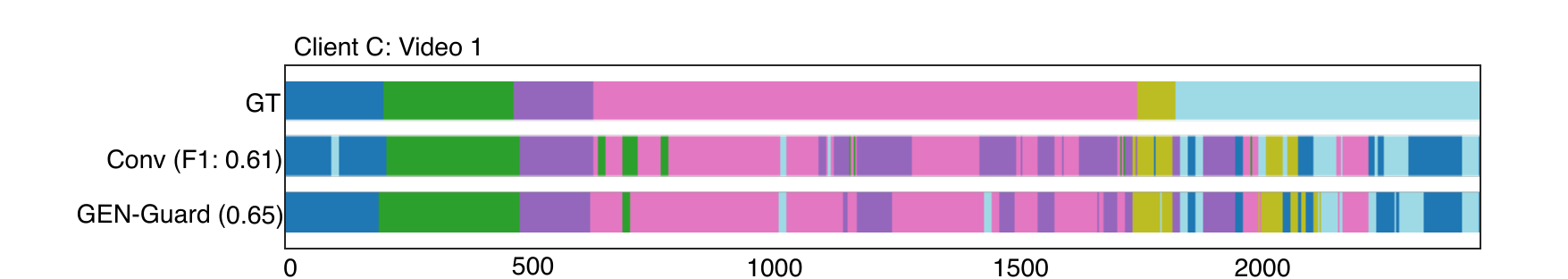}
        \vspace{-5.3mm}
    \end{subfigure}
    \begin{subfigure}{\linewidth}
        \centering
        \includegraphics[width=1\linewidth, trim={1mm 87mm 0 70mm}, clip]{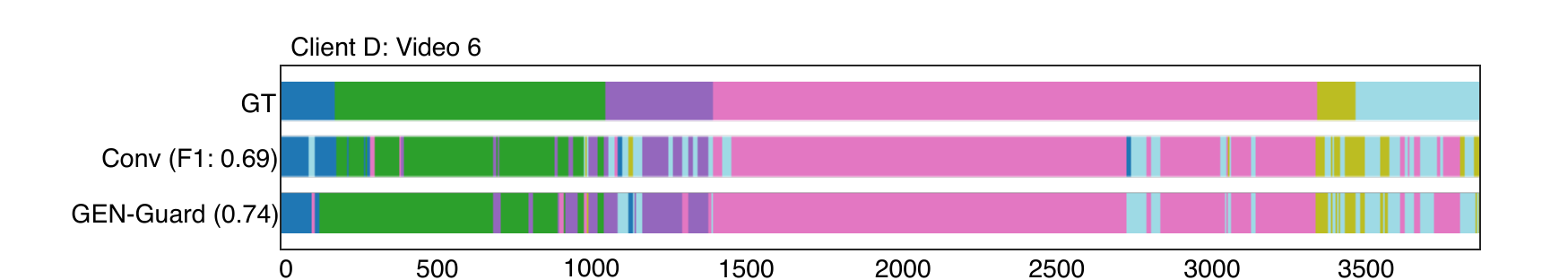}
        \vspace{-5.9mm}
    \end{subfigure}
    \begin{subfigure}{\linewidth}
        \centering
        \includegraphics[width=1\linewidth, trim={0 60mm 0 65mm}, clip]{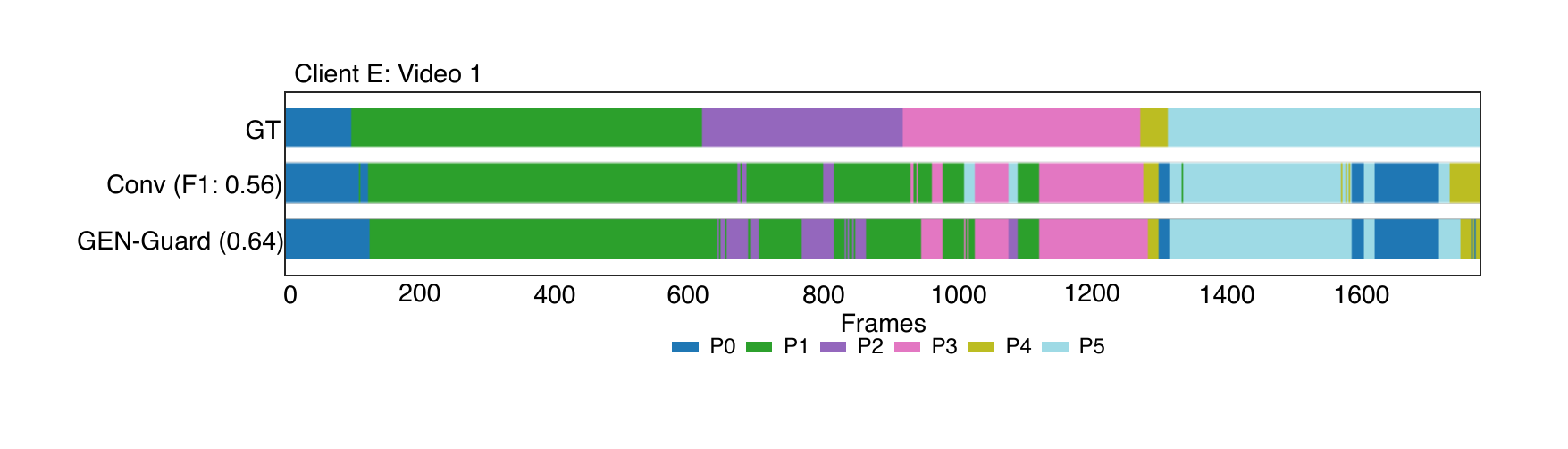}
        \vspace{2mm}
    \end{subfigure}
    \caption{Phase (P) recognition qualitative results across the Multi-Cholec clients. Each row displays the ground truth (GT), conventional (Conv), and GEN-Guard results on random test videos.}
    \label{fig:stacked_temporal_results}
\end{figure}

To qualitatively assess phase recognition, we visualize the surgical pipeline for five randomly selected test videos from different clients. Figure~\ref{fig:stacked_temporal_results} compares the ground truth with predictions from both the conventional and GEN-Guard methods. Overall, GEN-Guard shows fewer ``spiky'' misclassifications than the conventional baseline, especially during transitional phases.

Table~\ref{tab:polygen_gains_summary} shows similar trends on the PolypGen dataset. GEN-Guard consistently improves F1 (equivalent to Dice) across all metrics, with the largest gains in zero-shot generalization (Held-Out F1 up to 3.3 points, Out-Fed F1 up to 1.8 points). Even in worst-case scenarios, it substantially raises the minimum performance.

\begin{table*}[h]
\centering
\caption{GEN-Guard performance on PolypGen: average F1$_{\pm\text{std}}$ and worst-case correction.}
\label{tab:polygen_gains_summary}
\resizebox{400px}{!}{%
\begin{tabular}{lcccccc}
\toprule
\textbf{FL Protocol} & \multicolumn{2}{c}{\textbf{In-Fed F1 (Supervised)}} & \multicolumn{2}{c}{\textbf{Held-Out F1 (Zero-shot)}} & \multicolumn{2}{c}{\textbf{Out-Fed F1 (Zero-shot)}} \\
\cmidrule(lr){2-3} \cmidrule(lr){4-5} \cmidrule(lr){6-7}
& $\mathbf{F1}_{\text{CONV}}$ & $\mathbf{F1}_{\text{GEN-Guard}}$ ($\mathbf{\Delta}$) & $\mathbf{F1}_{\text{CONV}}$ & $\mathbf{F1}_{\text{GEN-Guard}}$ ($\mathbf{\Delta}$) & $\mathbf{F1}_{\text{CONV}}$ & $\mathbf{F1}_{\text{GEN-Guard}}$ ($\mathbf{\Delta}$) \\
\midrule
FedAvg   & $80.23_{\pm 0.06}$ & $\mathbf{81.63_{\pm0.04}}$ ($\mathbf{\uparrow 1.40}$) & $71.02_{\pm0.14}$ & $\mathbf{74.29_{\pm0.13}}$ ($\mathbf{\uparrow 3.27}$) & $84.07_{\pm0.03}$ & $\mathbf{85.64_{\pm0.04}}$ ($\mathbf{\uparrow 1.57}$) \\
FedProx  & $79.89_{\pm 0.04}$ & $\mathbf{82.04_{\pm0.04}}$ ($\mathbf{\uparrow 2.15}$) & $70.98_{\pm0.13}$ & $\mathbf{74.30_{\pm0.13}}$ ($\mathbf{\uparrow 3.32}$) & $83.56_{\pm0.03}$ & $\mathbf{84.79_{\pm0.04}}$ ($\mathbf{\uparrow 1.23}$) \\
SCAFFOLD & $80.92_{\pm0.05}$ & $\mathbf{82.85_{\pm0.05}}$ ($\mathbf{\uparrow 1.93}$) & $71.53_{\pm0.13}$ & $\mathbf{73.13_{\pm0.14}}$ ($\mathbf{\uparrow 1.60}$) & $84.16_{\pm0.04}$ & $\mathbf{86.01_{\pm0.04}}$ ($\mathbf{\uparrow 1.85}$) \\
\midrule
\multicolumn{7}{l}{\textbf{Worst-Case F1 Correction}} \\
\midrule
FedAvg   & $79.40$ & $\mathbf{81.82}$ ($\mathbf{\uparrow 2.42}$) & $79.16$ & $\mathbf{87.14}$ ($\mathbf{\uparrow 7.98}$) & $88.31$ & $\mathbf{90.94}$ ($\mathbf{\uparrow 2.63}$) \\
FedProx  & $86.30$ & $\mathbf{88.73}$ ($\mathbf{\uparrow 2.43}$) & $78.80$ & $\mathbf{81.76}$ ($\mathbf{\uparrow 2.96}$) & $79.95$ & $\mathbf{83.78}$ ($\mathbf{\uparrow 3.83}$) \\
SCAFFOLD & $85.54$ & $\mathbf{81.94}$ ($\mathbf{\uparrow 3.40}$) & $57.49$ & $\mathbf{60.83}$ ($\mathbf{\uparrow 3.34}$) & $78.34$ & $\mathbf{81.94}$ ($\mathbf{\uparrow 3.60}$) \\
\bottomrule
\end{tabular}%
}
\end{table*}

\begin{figure*}[t]
    \centering
    \begin{minipage}{0.47\textwidth}
        \centering
        \includegraphics[width=\linewidth, trim={40mm 30mm 50mm 30mm}, clip]{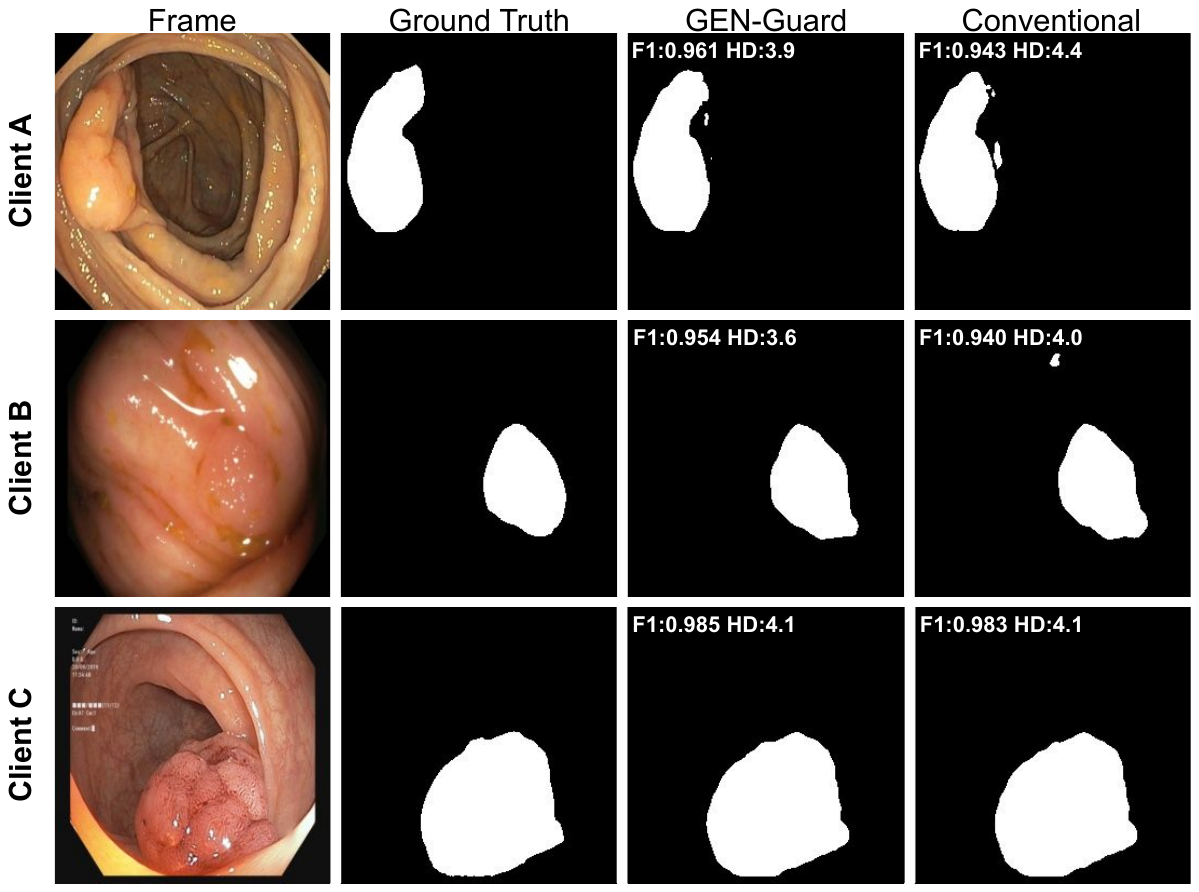}
    \end{minipage}
    \hfill
    \begin{minipage}{0.47\textwidth}
        \centering
        \includegraphics[width=\linewidth, trim={40mm 30mm 50mm 30mm}, clip]{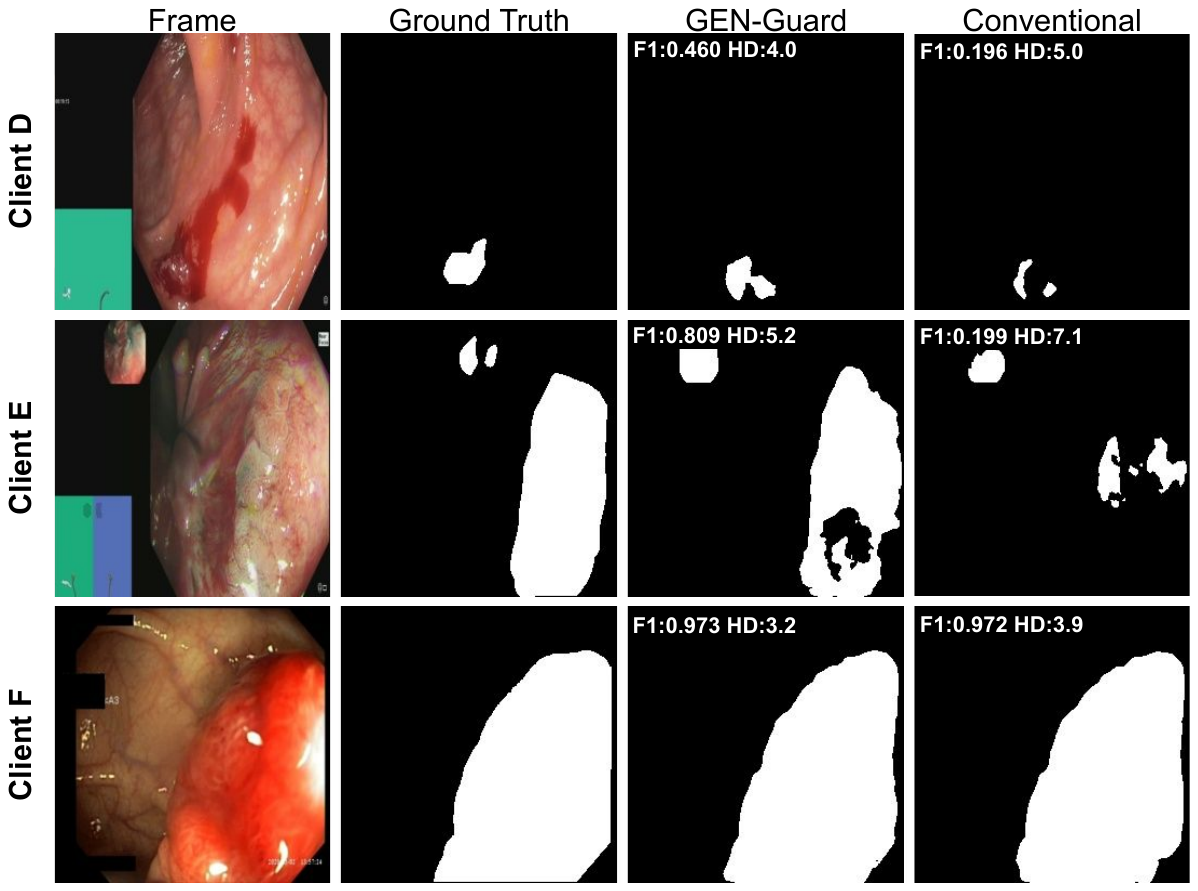}
    \end{minipage}
    \vspace{8pt}
    \caption{Segmentation qualitative results across the PolypGen clients. Each row compares the original frame, ground truth, conventional prediction, and GEN-Guard prediction on random frames from test videos.}
    \label{fig:combined_polyp_results}
\end{figure*}

Figure~\ref{fig:combined_polyp_results} presents a qualitative comparison of segmentation results across the PolypGen clients, with test samples randomly selected from each client. Overall, GEN-Guard enhances spatial robustness. Notably, when the $f_{\theta,\text{CONV}}$ and $f_{\theta,\text{CBE}}$ models exhibit high-confidence agreement, the GEN-Guard pipeline preserves the prediction. This behavior is clearly illustrated in a sample frame from Client C.

Overall, across both datasets, GEN-Guard consistently provides positive gains across all average metrics while reducing inter-institution variability, demonstrating its effectiveness in addressing Model Selection Failures (MSFs) and improving deployment reliability in federated surgical AI.

\subsection{Comparison with Personalized and Domain-Specific Baselines}

To address the challenges of data heterogeneity, we compare GEN-Guard against two established Personalized Federated Learning strategies as shown in Table~\ref{tab:master_comparison}: FedBN \citep{li2021fedbn}, which utilizes local batch normalization to capture site-specific statistics, and pFedMe \citep{t2020personalized}, which employs Moreau envelopes to balance global and local model updates.

Entries marked with an asterisk ($^*$) indicate models with local personalization, which requires an annotated local dataset for supervised fine-tuning from $f_{\theta,\text{CONV}}$ (up to $T_{\max}=5$ with early stopping). For FedBN, this adapts local BN statistics, while for pFedMe it corresponds to optimizing from the global proximal weight to a personalized local model.

In the Multi-Cholec task, personalized baselines such as FedBN achieve slightly higher Out-Fed performance, largely due to extensive supervised fine-tuning enabled by the large, expert-annotated Cholec80 dataset. In contrast, GEN-Guard achieves comparable Held-Out and strong Out-Fed robustness without any ground-truth labels. Unlike personalized baselines that rely on supervision, our method learns domain-invariant patterns via an unsupervised disagreement-aware signal.

\begin{table*}[h]
\centering
\caption{Personalized FL vs.\ proposed GEN-Guard: average F1$_{\pm\text{std}}$ across In-Federation (In-Fed), Held-Out, and Out-Federation (Out-Fed) clients. ($^*$) indicates methods with local personalization.}
\label{tab:master_comparison}
\resizebox{400px}{!}{%
\begin{tabular}{l ccc ccc}
\toprule
\multirow{2}{*}{\textbf{Method}} & \multicolumn{3}{c}{\textbf{Multi-Cholec}} & \multicolumn{3}{c}{\textbf{PolypGen}} \\
\cmidrule(lr){2-4} \cmidrule(lr){5-7}
& \textbf{In-Fed F1} & \textbf{Held-Out F1} & \textbf{Out-Fed F1} & \textbf{In-Fed F1} & \textbf{Held-Out F1} & \textbf{Out-Fed F1} \\
\midrule
\multicolumn{7}{l}{\textbf{Personalized Baselines:}} \\
FedBN  & $68.15_{\pm 1.84}$ & $67.29^*_{\pm3.11}$ & $\mathbf{68.85^*_{\pm5.19}}$ & $80.05_{\pm4.38}$ & $72.01^*_{\pm0.09}$ & $81.90^*_{\pm0.04}$ \\
pFedMe & $67.60_{\pm2.47}$ & $65.31^*_{\pm3.97}$ & $65.37^*_{\pm1.82}$          & $81.36_{\pm5.52}$   & $73.31^*_{\pm0.11}$ & $83.14^*_{\pm0.08}$ \\
\midrule
\multicolumn{7}{l}{\textbf{Proposed:}} \\
FedAvg+GEN-Guard   & $\mathbf{71.47_{\pm1.90}}$ & $66.89_{\pm4.27}$         & $67.18_{\pm1.40}$ & $81.63_{\pm0.04}$          & $74.29_{\pm0.13}$         & $85.64_{\pm1.57}$ \\
FedProx+GEN-Guard  & $67.31_{\pm1.48}$          & $67.72_{\pm3.83}$         & $64.27_{\pm1.20}$ & $82.04_{\pm0.04}$          & $\mathbf{74.30_{\pm0.13}}$ & $84.79_{\pm1.23}$ \\
SCAFFOLD+GEN-Guard & $71.25_{\pm2.60}$          & $\mathbf{68.79_{\pm1.86}}$ & $67.72_{\pm1.60}$ & $\mathbf{82.85_{\pm0.05}}$ & $73.13_{\pm0.14}$         & $\mathbf{86.01_{\pm0.04}}$ \\
\bottomrule
\end{tabular}%
}
\end{table*}

\subsection{Ablations}

To evaluate the contribution of each component, particularly in the DAD module, we performed a stepwise ablation study using FedAvg as the base FL algorithm (Table~\ref{tab:ablation_study_f1}). Since DAD operates on a discrete set of finalized candidate models, it is decoupled from the specific gradient descent mechanics or aggregation logic. FedAvg was chosen as a representative baseline, as it underlies more advanced federated methods such as FedProx and SCAFFOLD.

\begin{table*}[h]
\centering
\caption{Ablation study (F1-Score) across two datasets. ``Unseen'' denotes the mean of Held-Out and Out-Fed F1 scores, with $\pm$Std between these two values.}
\label{tab:ablation_study_f1}
\resizebox{400px}{!}{%
\begin{tabular}{l c c c c}
\toprule
\textbf{Ablation Model / Component Added} & \multicolumn{2}{c}{\textbf{Multi-Cholec F1}} & \multicolumn{2}{c}{\textbf{PolypGen F1}} \\
\cmidrule(lr){2-3} \cmidrule(lr){4-5}
& \small{In-Fed (Supervised)} & \small{Unseen (Zero-shot)} & \small{In-Fed (Supervised)} & \small{Unseen (Zero-shot)} \\
\midrule
$\hat{f}_{\theta, \text{CONV}}$                                                  & $69.53_{\pm2.14}$ & $65.20_{\pm0.62}$ & $80.23_{\pm0.06}$ & $77.54_{\pm6.52}$ \\
\midrule
Conventional Distillation ($\mathcal{L}_{\text{KL}}$)                           & $69.75_{\pm2.15}$ & $65.45_{\pm0.53}$ & $80.40_{\pm0.06}$ & $77.80_{\pm6.45}$ \\
\quad + Disagreement Split (Uniform $w$)                                         & $70.15_{\pm2.08}$ & $66.00_{\pm0.35}$ & $80.42_{\pm0.05}$ & $78.40_{\pm6.42}$ \\
\quad + Disagreement Focus ($w_{\text{disagree}} > w_{\text{agree}}$)            & $70.05_{\pm2.01}$ & $65.51_{\pm0.30}$ & $80.95_{\pm0.06}$ & $78.85_{\pm6.22}$ \\
\quad + Cosine Similarity ($\mathcal{L}_{\text{COS}}$)                           & $70.81_{\pm1.95}$ & $66.75_{\pm0.21}$ & $81.19_{\pm0.04}$ & $79.04_{\pm5.97}$ \\
\quad + Temperature Differential ($T_{\text{disagree}} \ne T_{\text{agree}}$)   & $71.29_{\pm1.92}$ & $66.98_{\pm0.19}$ & $81.49_{\pm0.04}$ & $79.50_{\pm5.95}$ \\
\midrule
\textbf{Final Model (GEN-Guard)}                                                 & $\mathbf{71.47_{\pm1.90}}$ & $\mathbf{67.03_{\pm0.14}}$ & $\mathbf{81.63_{\pm0.04}}$ & $\mathbf{79.96_{\pm5.67}}$ \\
\bottomrule
\end{tabular}%
}
\end{table*}

The ablation studies show that starting from $\hat f_{\theta, \text{CONV}}$, standard distillation offers minimal improvement. Adding disagreement-weighting and cosine similarity increases both in-federation and zero-shot F1 scores. Overall, the full GEN-Guard model achieves the best results.

\begin{figure*}[t]
    \centering
    \begin{minipage}{0.49\textwidth}
        \centering
        \includegraphics[width=\linewidth]{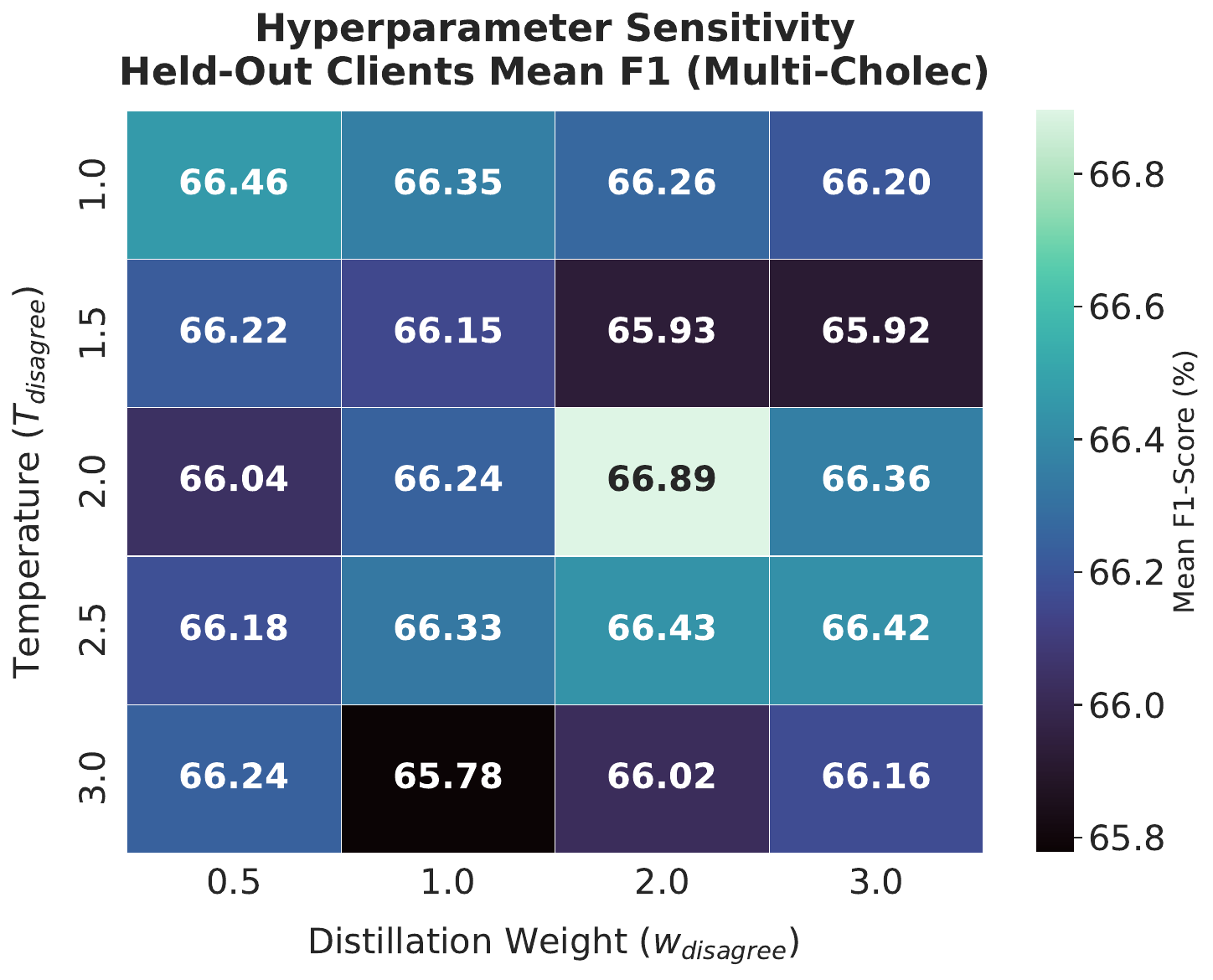}
    \end{minipage}
    \hfill
    \begin{minipage}{0.49\textwidth}
        \centering
        \includegraphics[width=\linewidth]{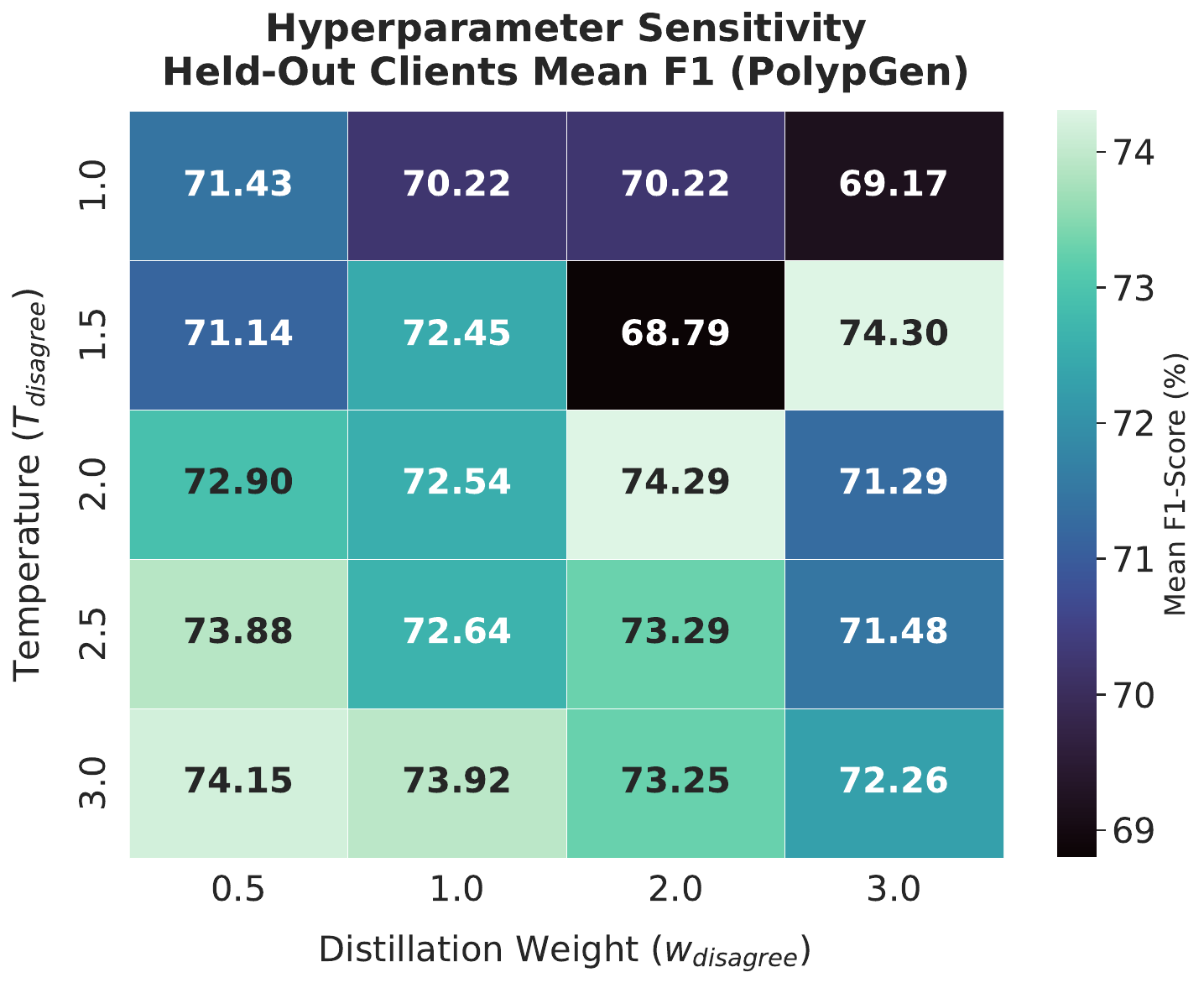}
    \end{minipage}
    \vspace{8pt}
    \caption{Heatmaps visualizing average F1$_{\pm\text{std}}$ across Held-Out clients.}
    \label{fig:combined_sensitivity}
\end{figure*}

We conducted a comprehensive grid search over the temperature $T_{\text{disagree}} \in [1.0, 3.0]$ and distillation weight $w_{\text{disagree}} \in [0.5, 3.0]$ (Figure~\ref{fig:combined_sensitivity}). The Agreement parameters were fixed at $T_{\text{agree}} = 1.0$ and $w_{\text{agree}} = 1.0$, reflecting standard Knowledge Distillation (KD). This study was performed on Held-Out clients to assess true zero-shot generalization to unseen clinical distributions. FedAvg was used as the base algorithm for two reasons: (1) as the foundational and widely adopted federated learning protocol; and (2) because of the post-hoc nature, the disagreement-aware logic is decoupled from the training optimizer's gradients.

Overall, both phase recognition (left) and spatial segmentation (right) show robust performance across parameter variations, with a stable plateau near the chosen configuration ($T_{\text{dis}} = 2, w_{\text{dis}} = 2$).

\subsection{Computational Footprint}

We evaluate the computational footprint of GEN-Guard across two datasets using a single NVIDIA V100 per client. Table~\ref{tab:compact_efficiency} reports the mean and standard deviation over all federated clients.

In the Multi-Cholec phase recognition task, the detection phase (CBE) achieves a throughput of 120.30~FPS, about five times faster than the training loop. The DAD correction phase remains efficient (39.16~FPS), and the distilled model sustains high inference speeds suitable for real-time surgical deployment.

For the PolypGen segmentation task, while standard training exceeds 2.6~GB of VRAM, the DAD correction phase reduces peak usage to 489.37~MB. The distilled inference model remains equally lightweight for resource-constrained environments.

\begin{table}[ht]
\centering
\caption{Comparison of computational efficiency across two datasets.}
\label{tab:compact_efficiency}
\resizebox{\columnwidth}{!}{%
\begin{tabular}{llcc}
\toprule
\textbf{Dataset} & \textbf{Step} & \textbf{Throughput (FPS) $\uparrow$} & \textbf{Peak VRAM (MB) $\downarrow$} \\
\midrule
\multirow{4}{*}{\textbf{Multi-Cholec}}
& FL Training    & $23.18_{\pm 10.1}$ & $1091.65_{\pm 25.6}$ \\
& CBE Detection  & $120.30_{\pm 20.5}$ & $580.11_{\pm 1.0}$ \\
& DAD Correction & $39.16_{\pm 9.7}$  & $489.37_{\pm 12.0}$ \\
& DAD Inference  & $116.30_{\pm 52.4}$ & $466.89_{\pm 0.9}$ \\
\midrule
\multirow{4}{*}{\textbf{PolypGen}}
& FL Training    & $10.42_{\pm 3.2}$  & $2643.54_{\pm 0.1}$ \\
& CBE Detection  & $13.80_{\pm 4.1}$  & $577.15_{\pm 2.4}$ \\
& DAD Correction & $15.13_{\pm 4.7}$  & $453.13_{\pm 0.3}$ \\
& DAD Inference  & $19.21_{\pm 8.7}$  & $419.75_{\pm 1.2}$ \\
\bottomrule
\end{tabular}%
}
\end{table}

\section{Discussion}

GEN-Guard detects MSFs, but its precision depends on the representativeness of client validation sets during CBE. Noisy or unrepresentative validation data may lead to a suboptimal choice of $f_{\theta,\text{CBE}}$, potentially affecting the subsequent DAD correction.

\textit{Limitations:}
Nevertheless, the disagreement-aware DAD acts as a protective buffer by applying corrections only where models diverge, thereby limiting the propagation of errors from an imperfect $f_{\theta,\text{CBE}}$ and preventing catastrophic generalization collapse. Empirically, the framework maintains stable performance even under imperfect model selection. For example, evaluations on highly divergent datasets, such as Cholec80 (France) versus Italian surgical centers, indicate that GEN-Guard can still extract domain-invariant patterns. Similar robustness is observed under extreme data imbalance in Multi-Cholec, where the method remains effective across both the smallest and largest out-of-federation cohorts (Clients E and B), a trend also reproduced in the PolypGen segmentation task (Clients F and C). Nonetheless, highly biased validation splits may still reduce MSF detection sensitivity.

\textit{Clinical implications:}
Performance leakage poses potential patient safety risks when federated models are deployed across institutions with unseen procedural styles, imaging systems, or patient populations. Even moderate degradation in phase recognition or polyp segmentation may affect downstream clinical workflows, including intraoperative guidance, documentation automation, and quality monitoring. By explicitly detecting and correcting MSFs prior to deployment, GEN-Guard introduces an additional validation safeguard that can improve the reliability of cross-institutional surgical AI systems.

\textit{Ethical considerations:}
Residual performance leakage may also raise fairness concerns if models systematically underperform on certain institutions or patient subgroups. While GEN-Guard reduces this risk, it cannot eliminate it entirely, as unseen distributions may still differ substantially from those observed during training. Responsible deployment therefore requires continuous post-deployment monitoring, transparent reporting of cross-institutional variability, and appropriate human oversight.

Future work will investigate strategies to further improve scalability and robustness, such as automated weighting of validation clients based on domain diversity or performing CBE on a subset of anchor clients to reduce computational overhead in large federations while preserving corrective capability.

\section{Conclusion}

This work presents GEN-Guard, a post-hoc deployment-oriented federated learning framework that detects and corrects Model Selection Failures (MSFs), enhancing cross-institutional generalization in surgical AI. MSFs were observed in over 80\% of experiments, emphasizing the importance of bias correction and robust model selection. GEN-Guard improves in-federation F1 scores by up to 2 points, zero-shot generalization F1 on unseen clients by up to 3 points, and worst-case F1 by 3--9 points, all with minimal computation and no extra communication during the main federated training. These results demonstrate GEN-Guard's practicality and effectiveness for real-world deployment of federated surgical AI.

\section{Compliance with Ethical Standards}

\textbf{Ethical approval:} This article does not contain any studies with human participants or animals performed by any of the authors.

\textbf{Competing interests:} The authors declare no conflict of interest.

\textbf{Informed consent:} This manuscript does not contain any patient data.


\section{Disclosures}
Pietro Mascagni and Nicolas Padoy are co-founders and
shareholders of Scialytics. The other co-authors do not have
any conflict of interests to disclose.

\section{Acknowledgments}

This work was supported by French state funds managed within the Plan Investissements d'Avenir by the ANR under references ANR-22-FAI1-0001 (project DAIOR) and ANR-10-IAHU-02 (IHU Strasbourg).

\bibliographystyle{model2-names}
\bibliography{sn-bibliography}

\end{document}